# Application of Yolo on Mask Detection Task

Ren Liu
Georgia Institute of Technology
Atlanta, Georgia, United States
Author's Email: rliu384@gatech.edu

Ziang Ren
Georgia Institute of Technology
Atlanta, Georgia, United States
Author's Email:zren66@gatech.edu

*Abstract*—2020 has been a year marked by the COVID-19 pandemic. This event has caused disruptions to many aspects of normal life. An important aspect in reducing the impact of the pandemic is to control its spread. Studies have shown that one effective method in reducing the transmission of COVID-19 is to wear masks. Strict mask-wearing policies have been met with not only public sensation but also practical difficulty. We cannot hope to manually check if everyone on a street is wearing a mask properly. Existing technology to help automate mask checking uses deep learning models on real-time surveillance camera footages. The current dominant method to perform real-time mask detection uses Mask-R-CNN with ResNet as backbone. While giving good detection results, this method is computationally intensive and its efficiency in real-time face mask detection is not ideal. Our research proposes a new approach to the mask detection by replacing Mask-R-CNN with a more efficient model "YOLO" to increase the processing speed of real-time mask detection and not compromise on accuracy. Besides, given the small volume as well as extreme imbalance of the mask detection datasets, we adopt a latest progress made in few-shot visual classification, simple CNAPs, to improve the classification performance.

## I. Introduction

Under the current COVID-19 background, it is vitally important to control the spread of the disease. Studies have shown that mask-wearing can significantly decrease the risk of COVID-19 transmission. However, it is unreasonable to expect that everyone is able and willing to wear a mask. Many countries and regions have imposed laws to enforce mask wearing in public places. However, these legislations are incredibly difficult to enforce using pure human labor, especially in public places. Automation techniques are necessary to perform detections in real time. There are existing methods that automates mask-detection.

The most dominant of these methods employ a deep neural network model called Mask-RCNN to automatically detect people who are wearing and not wearing masks from a video source. The problem with these methods is that Mask-RCNN is computationally heavy. This means that systems that rely on Mask-RCNN requires adequate computing power to run the model in real time. The bar in computing power presents a difficulty to deploy real-time mask detection systems, especially in places where such devices are not affordable. Our project aims to reduce the computation cost of such automated mask detection system by replacing the Mask-RCNN with the YOLO model, which achieves the same level of accuracy and is two-degrees faster. This significant reduction in computation costs allows for less expensive device to be deployed for mask detection. We believe that this would help in enforcing mask wearing policies and consequently reducing the cases of COVID-19 infection.

## II. Related Works

R-CNN and Mask-R-CNN is a family of convolutional neural networks (CNNs) that perform object detection on images. Proposed in 2014, the original R-CNN is the first to use CNNs to perform object detection. R-CNNs yields significantly better results compared to traditional feature extraction methods such as Scale-invariant feature transform (SIFT), Histogram-of-oriented-gradients (HoG) features and bag-of-features methods [3]. R-CNN detects objects by making around 2000 region proposals from the input image via the Selective Search algorithm, warping the regions in the same way, and passing them through a CNN feature extractor. The output of the CNN is then fed into a simple Support Vector Machine (SVM) classifier for classifying objects in the proposed regions and a regressor for generating the bounding boxes. R-CNN is extremely slow because detecting objects in a single image requires passing all of the region proposals through the CNN. Running time for R-CNN predictions is usually around 1 minute. The R-CNN is very difficult to train because it is RAM heavy and requires training the CNN, the SVM and the regressor. [3]. Improvements to R-CNN was swiftly made with its original author introducing Fast R-CNN in 2015 [2]. Fast R-CNN achieves significantly better running time by eliminating the repeated feed-forward feature extraction of proposed regions. This is done by directly feeding the image through a CNN and use the resulting feature map for region proposals instead via Selected Search. The proposals were then warped into the same shape through a special Region-of-Interest (RoI) layer and then fed into fully connected (FC) layers. The SVM is replaced by a SoftMax layer to perform classification, hence making Fast R-CNN easier to train. Ren et al. further improved on Fast R-CNN by replacing the Selective Search algorithm, which turned out to be the bottle-neck in speed and accuracy of R-CNNs, with a CNN. This CNN, called the Region-proposal Network (RPN) is responsible for proposing regions instead. RPN works in accordance with RoI pooling by sharing the feature maps generated by the feature extractor CNN. Result is a model that can be run in real-time at 5 frames per second on a graphical processing unit (GPU) [9]. Adding arguments on why this is still not okay, we want a faster algorithm: lower computation cost so that we can process more data on cheaper machines/on edge devices to improve coverage of this automated face-mask detection system.

YOLO, abbreviation for "You only look once", is another type of model that performs object detection on images. Instead of taking the two-stage propose-and-classify approach defined by the family of R-CNN models, YOLO achieves object detection in a single stage by treating the task as a regression problem. This allowed YOLO to achieve a 45 frames-per-second detection speed on a Titan X GPU [7]. Proposed in 2016 by Redmond et al., the original YOLO model divides the input image into a 19×19 grid of cells. Each cell is responsible for predicting 2 bounding boxes by giving a confidence level on if an object exists in each bounding box. The bounding box predictions are non-max suppressed. The final prediction results are those with the highest Intersection over Union (IoU) scores [7]. The original YOLO model had several limitations. 1) Difficulty to detect objects that appear in groups or have unusual aspect ratios or configurations, because each grid cell is responsible for predicting only 2 bounding boxes of fixed aspect ratios; 2) Localization inaccuracies because the loss function treats low and high dimension localization errors equally. Redmon et al. made improvements on the YOLO model in their subsequent research by adding Batch Normalization layers, replacing fully-connected layers with anchor boxes, and using a custom DarkNet-19 convolutional layer to reduce the number of floating-point observations. Redmon et al. made some final consolidations in YOLOv3, the most notable of which is expanding the DarkNet-19 to a larger but more accurate DarkNet-53 CNN that incorporates residual layers while adding minimal computational overhead. The result is a fast, accurate model that can process images at 30 frames-per-second on a Pascal Titan X GPU[8]. Notably, YOLO provides different model architectures that allows for different trade-offs in processing speed and accuracy.

Few-Shot Visual Classification. The goal of few-shot learning is to automatically adapt models such that they work well on instances from classes hardly seen at training time. Most of last decade's few-shot learning works can be differentiated along two main axes: 1) how images are transformed into vectorized embeddings, and 2) how" distances" are computed between vectors in order to assign labels.

Siamese network [5], an early approach to few-shot learning and classification, used a shared feature extractor to produce embeddings for both the support and query images. Relation networks, and recent GCNN [4] variants, extended this by parameterizing and learning the classification metric using a Multi-Layer Perceptron (MLP). Matching networks [11] learned distinct feature extractors for support and query images which were then used to compute cosine similarities for classification. The feature extractors used in such works are usually obtained by fine-tuned transfer-learned networks.

This paper adopts Simple CNAPS to solve problems caused by small and imbalanced dataset. Methods that are most similar to approach are Simple CNAPS [1], CNAPS [10] (and the related TADAM) and Prototypical networks. It is the state-of-the-art approach for few-shot image classification. In this kind of methods, CNAPS use a pretrained feature extractor augmented with FiLM layers [6] that are adapted for each task using the support images specific to that task. Different from CNAPS, simple CNAPS [1] makes improvements in the following ways. CNAPS has demonstrated the importance of adapting the feature extractor to a specific task. Simple CNAPS demonstrates an improved choice of Bregman divergence and proposes that the use of the Mahalanobis distance is helpful for image classification. In addition, simple CNAPS makes it no extra parameters, which is easier for training and can be used for different classification tasks.

### III. IMPLEMENTATION

*A. Dataset*

Masks play a significant role in protecting the health of individuals against virus spread in air, as is one of the few precautions available for COVID-19 in the absence of immunization. Hence, it is very important for us to detect whether an individual wear a mask and whether they wear correctly as a means of tracing the infection.

Currently, data-driven detection and classification models must be fitted with a dataset to function properly. Mask detection and classification dataset in this paper come from one of the latest Face Mask Detection and Kaggle.

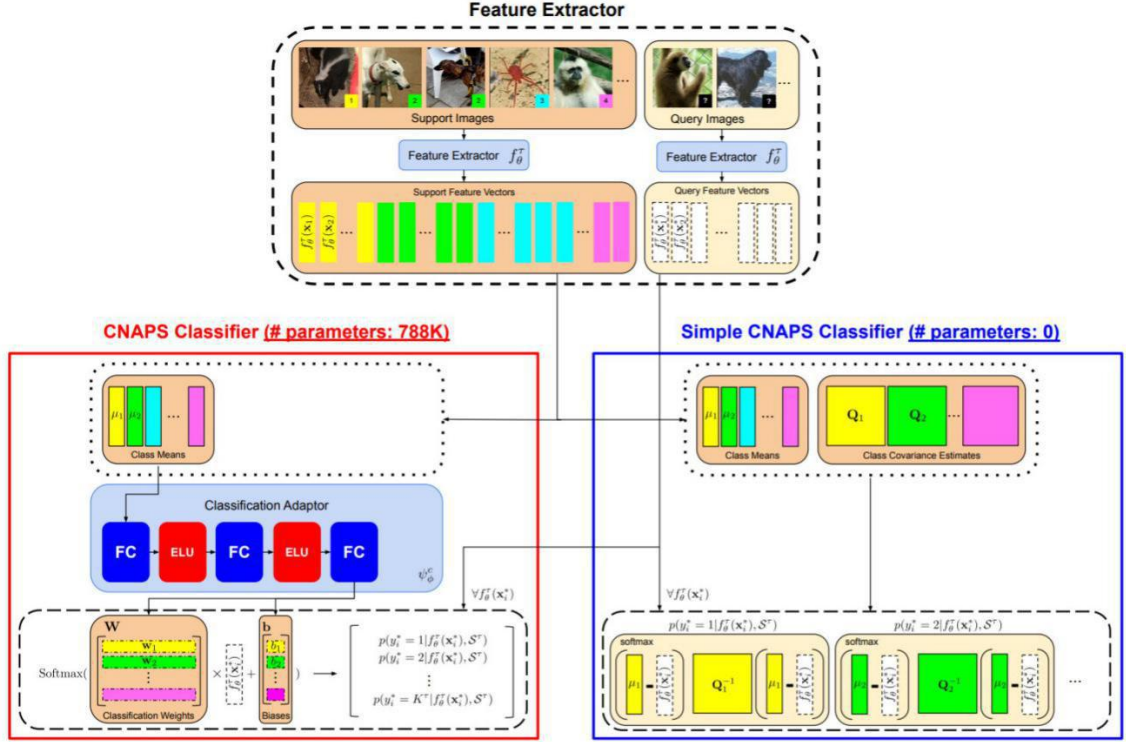

Figure 1. Structures of simple CNAPS [1]

This dataset contains 853 images belonging to the 3 classes described below, as well as their bounding boxes in PASCAL VOC format:
- With mask
- Without mask
- Mask worn incorrectly

This dataset is well prepared for detection and classification models, that is, in every single image, there might be multiple targets with different classes. This task is what Yolo framework designed for.

Additionally, based on this dataset, we also built a simpler dataset consisting of target slices in the original images, in order to train and test Yolo-based classification-only models. In the training set, there are 3145 images, with 2546 with mask, 508 without mask, and 91 masks worn incorrectly. The above numbers tell us that the dataset is limited in size and is very biased towards the "Wearing Mask" class.

*B. Traditional Detection & Classification Task*

While most proposed solutions in the Kaggle Face Mask Detection Dataset competition use Faster R-CNN to achieve good detection results, these R-CNN models are still computationally expensive and does not have good frames-per-second performance in real time. Our first task is to adjust a well-trained Yolov3 model to this traditional detection & classification task to obtain equal accuracy and faster speed than the proposed solutions.

Our work began with the GitHub repository ultralytics/yolov3. The repository is maintained by ultralytics, a U.S.-based particle physics and AI startup with over 6 years of expertise supporting government, academic and business clients.

Based on their work and our original dataset, we did the following training and tuning strategies:

- **Weighted Loss Function:** We found that, during training, the model has a tendency to converge better in bounding box and object loss and less well in class loss. To mitigate this problem, we used a weighted loss function that can be expressed as:

$$L = \alpha L_{cls} + \beta L_{obj} + (3 - \alpha - \beta) L_{bbox} \quad (1)$$

- **Data Augmentation:** based on the limited size of training data, we adopted a series of data augmentation methods including:
  - Random Rotation / Translation / Scaling / Flip
  - Random Hue / Brightness / Inverse / Saturation
  - Random Gaussian Blur
- **Undersampling:** Given that target *Correctly wearing the mask* has much more samples than the other two classes, we believe that it is necessary to adopt undersampling technique to make the training data more balanced.
- **Hyper-parameters tuning:** We manually tuned the hyperparameters including batch size, grid size, learning rate schedule.

## C. Few-Shot Classification-Only Task

While fitting the traditional detection & classification model, we found that our model had a low classification accuracy. We decided to investigate the cause. After we built the classification-only dataset, we found that the dataset is imbalanced in the distribution of the class samples. Images that belong to Mask worn incorrectly class are very few. For this reason, we adopted the latest Few-Shot classification techniques to perform this classification only task. Goals are:

- To get better classification scores than the traditional detection & classification models on the target slices
- To compare the performance of feature extraction between two backbones that have relatively equal number of parameters: Mobilenetv3 backbone and YoloNano

To achieve goals, we implemented Simple CNAPS. Simple CNAPS [1] is first proposed by Peyman Bateni, et al., based on the previous work about CNAPS, which is one of the state-of-the-art in the Few-Shot classification domain. Compared with the original CNAPS, simple CNAPS makes the few-shot classifier no parameters. Their work is published in CVPR 2020 and showed that simple CNAPS can achieve good performance in few-shot classification task that is even better than the conventional CNAPS. Because there is no source code that has implemented simple CNAPS, we reproduce the model ourselves.

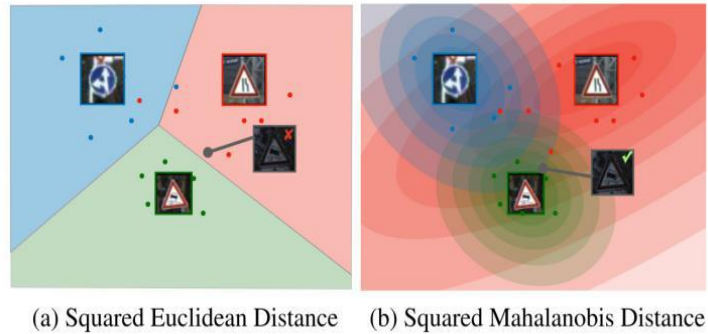

Figure 2. Difference between Euclidean Distance and Mahalanobis Distance

One of the great differences between simple CNAPS classifier and traditional FC-layer Classifier used in CNN, is that simple CNAPS adopts the concept of Mahalanobis distance (class-covariance based) instead of Euclidean distance as the metric. An advantage of using a class-covariance based metric during classification is that such metrics take the distribution in feature space into account and give improved decision boundaries for non-linear classifiers. In Mask Detection task, the decision boundary between With mask and Mask worn incorrectly are very close. We identify this as the source of misclassification by our model.

Additionally, in the implementation of simple CNAPS, every class is treated equally with its Class Mean and Class Covariance Estimates, which indicates that this classifier will be less influenced by the imbalanced dataset.

Furthermore, adopting simple CNAPS decreases the number of FC layers with only a few extra parameters stored. This result is also in line with our initial purpose of the this research: to train a high efficiency detection and classification model for real-time use cases.

To implement the simple CNAPS-based classifier, we designed a full 3-stage training-inferring process by using simple CNAPS for classification. This is not mentioned in the original paper. The 3-stage process includes pretraining, finetuning and querying(Figure 2).

## D. Video Processing

We use OpenCV visualize the prediction results in videos. OpenCV supports reading streams of videos from external devices and files from the local file system. Given a trained model on a mask-detection dataset, we expect the output of the model to contain at least the following fields:

- An array of images used in the prediction
- An array of predictions generated by the model, of tuples of the following format
    (a) x, y coordinates of the top left corner of the bounding box, normalized to image width and height.
    (b) x, y coordinates of the bottom right corner of the bounding box, normalized to image width and height.
    (c) a floating-point confidence level
    (d) a number indicating the predicted class 3. an array of label names

The video source is read as an iterable stream of frames of images. Each frame of image is passed into our model at their original height and width (e.g., 1080 pixels wide, 1920 pixels high). Our model generates inference results conforming to the above format. We use the results to draw the bounding boxes, predicting class names and confidence level for each detected object (face, face masks, face masks worn incorrectly) on this frame of image. The drawn frame is then passed into a video encoder to be saved as a frame in the output video. The end result is a new video with the above visualizations with MPEG-4 encoding. The input video is not modified in any way.

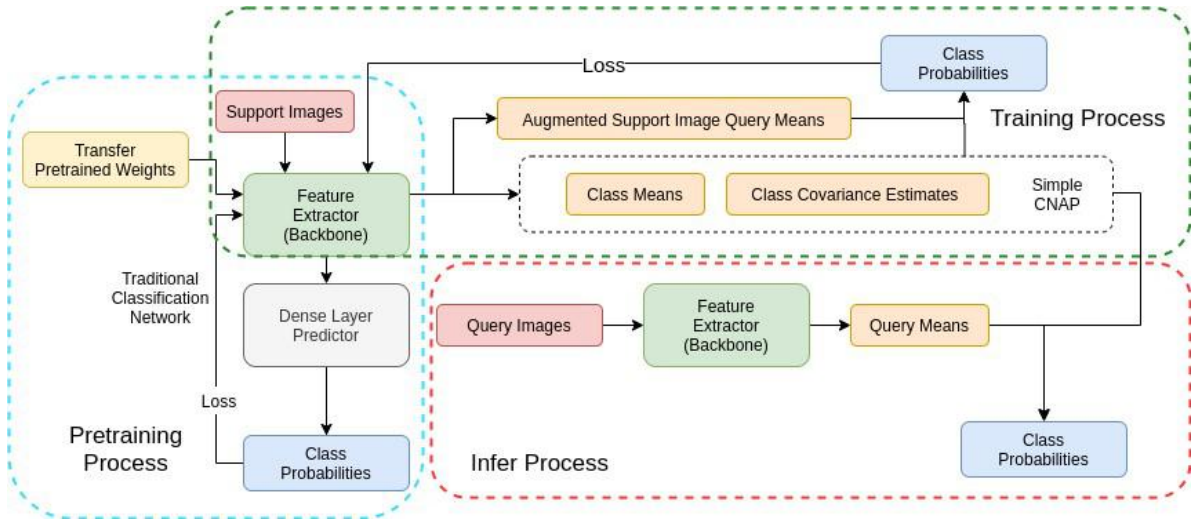

Figure 3. Designed training routine for simple CNAPS.

Processing videos with OpenCV adds overhead to model prediction. The overhead comes from reading frames from the input video, drawing the visualizations and writing the drawn frame to the output video. Model is very performant, achieving a 2 frames-per-second on a modest dual-core Intel Xeon CPU at 1920×1080 resolution.

## IV. RESULTS

### A. Traditional Detection & Classification Task

We trained a Faster RCNN, an original Yolov3 and our customized weight loss Yolov3 with fine-tuned hyper-parameters, based on original Face Mask detection dataset. The original data set is randomly divided into training set and validation set in a ratio of 4:1. The training process was carried on a server with modest dual-core Intel Xeon CPU and NVidia RTX 2080 Ti video card. Each model is trained in an 8-batch size with 200 epochs. The following table shows their performances on the same validation set. In the weighted loss Yolov3, we set:

$$\alpha = 1.25, \beta = 1.0 \qquad (2)$$

From the results above, we see clearly that Faster RCNN tends to perform better than Yolov3. We believe that it might because the Faster R-CNN uses a larger feature extractor (ResNet-50). However, with a little loss in the accuracy, Yolov3 has a much better running speed.

TABLE I. PERFORMANCES OF DIFFERENT MODELS

| Method | Precision | F1 | Speed* |
|---|---|---|---|
| Faster R-CNN | 0.932 | 0.721 | 175.2 |
| Yolov3 | 0.905 | 0.706 | 96.42 |
| Yolov3 Weighted Loss | 0.919 | 0.731 | 95.78 |

*Here the speed is calculated in the time cost (ms) per 100 images

We can also see that the F1 score is not as good as the precision. This reveals the imbalance of the dataset as we have mentioned before. However, as we applied our customized weighted loss, which added the proportion of $L_{cls}$, our new model achieves a better F1 score. The experiment results proved our hypothesis.

### B. Few-Shot Classification-only Task

We trained the traditional classification network with FC-layer predictor. We chose a Mobilenetv3 backbone with 2,826,736 trainable variables and a YoloNano backbone with 2,771,991 trainable variables as our candidate feature extractors. Experiment results are listed as follows

In order to avoid the influence of imbalanced dataset, we adjust the ratio of each class in the validation set. Our validation set has 183 "With Mask" images, 40 "Mask worn incorrectly" images, and 129 "No Mask" images.

From the table above we see the feature extracting ability of YoloNano is not worse than the state of art mobile feature extractor MobileNetv3 (large version). What's more, YoloNano shows a better performance in avoiding overfitting on small datasets.

TABLE II. FEW-SHOT CLASSIFICATION TRAIN SET

| Settings | Mobilenetv3 | YoloNano |
|---|---|---|
| FC Classifier | 0.9014 | 0.8693 |
| Simple CNAPS-50 | 0.8243 | 0.8245 |
| Simple CNAPS-100 | 0.8718 | 0.8752 |
| Simple CNAPS-500 | 0.8574 | 0.8457 |
| Simple CNAPS-full | 0.8693 | 0.8543 |

*Here "Simple CNAPS-50" means that we set the size of support images of CNAPS as 50, and "Simple CNAPS-full" means we set all images in the training set as support images when querying.

TABLE III. FEW-SHOT CLASSIFICATION VALIDATION SET

| Settings | Mobilenetv3 | YoloNano |
|---|---|---|
| FC Classifier | 0.8457 | 0.8590 |
| Simple CNAPS-50 | 0.8536 | 0.8557 |
| Simple CNAPS-100 | 0.8617 | 0.8876 |
| Simple CNAPS-500 | 0.8777 | 0.8776 |
| Simple CNAPS-full | 0.8457 | 0.8513 |

Besides, we see a remarkable improvement in test accuracy after using simple CNAPS instead of the FC layer predictor.

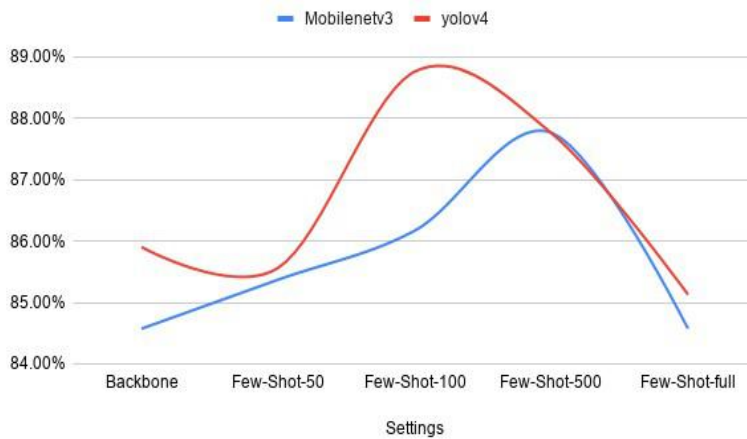

Figure 4. Curves of validation set performance

When inspecting the curves, we see that when the support images number increase, the simple CNAPS's performance first improves and then worsens. This matched the results that Peyman et al. mentioned in their paper. Hence, when using simple CNAPS, tuning the number of support images is be very important.

## V. DISCUSSION

### A. Performance Evaluation for the Yolo Family

When Faster RCNN and Mask RCNN are so popular, why we still consider networks from the Yolo family. With this question, our work showed that the Yolo family networks have great capabilities in feature extraction and object detection that are no worse than the current state-of-the-art such as faster/mask RCNN and MobileNet. Additionally, Yolo family networks have faster execution times, which is better-suited for practical application in real-time use cases that we proposed: detecting masks in real time.

### B. Using Simple CNAPS in Practical Task

We first designed a full training process for simple CNAPS in a real-life task, and proved that it is helpful in improving the few-shot class classification scores. The next step might be adopting simple CNAPS on the traditional detection & classification frame works.

### C. Improvements on Video Processing Pipeline

Further improvements can still be made based on our video processing pipeline. These improvements may not be directly related to our model. 1) We do not need to pass every frame to our model. This is based on the assumption that continuous frames in the input video are likely to contain similar content. By strategically skipping frames, we can improve the processing speed of our entire pipeline. 2) We can incorporate instance tracking in our video processing pipeline. Instance tracking compares the pixel content of frames to predict the movement of detected objects. This takes significantly less floating-point operations than passing frames to our model for inferencing and hence will improve our overall processing speed.

### D. Meaning of Project

This project has practical value under the current context of the COVID-19 pandemic. Pipeline is already capable of detecting people with, without and incorrectly wearing masks with reasonable accuracy. With some improvements, we envision that product can be used as a component in a contact tracing system. Product is also relatively computationally efficient. The hardware threshold for deploying our project is low. This means that product is less restricted by budget or the level of economic development at the location of its deployment and hence can reach more places where COVID-19 infections pose more threat to people.

*E. Privacy Concerns*

Deep learning models have vulnerabilities. While it is possible to conduct adversarial attacks on our model if it is deployed, such attacks are unlikely not cause direct, physical harm to people whose faces are detected. It is worth mentioning that, with minimum improvements, our model is capable of memorizing detected faces (e.g., through a face-recognition deep-learning framework). This is a likely use case if our model is incorporated into a contact-tracking system where facial-recognition and storing faces are required. Facial features are generally considered to have some level of privacy. In such cases, we should implement counter measures such as implementing safe deep learning models, obfuscating stored faces and putting our product behind a safe strong-point to protect the stored human faces.

## VI. CONCLUSION

In this paper, we examined the performance of neural networks from the Yolo family in real-time detection tasks. Results showed that this Yolo is capable of achieving state-of-the-art performance in object detection and classification with a much lower inference time. This showed that Yolo is well-suited for detection and classification tasks in real-time settings such as the one we proposed: face-mask detection in real time. Additionally, we implemented the Simple CNAPS model and found that it improves our model performance on the small and biased dataset that is available to us. We have also implemented a sample video processing pipeline to demonstrate our model performance.

## VII. APPENDIX

*A. Dataset*

As detailed in section 3.1, our dataset is acquired from Kaggle. The dataset contains 853 images and one boxing box specification file in PASCAL VOC format for each image. This dataset is used to build simpler dataset containing only the target slices of the original images. This dataset is used to train and test the Yolo-based classification-only models. There are 3145 images in the training set. 2546 with mask, 508 without mask, 91 masks are worn incorrectly. This dataset is very small and imbalanced.

*B. Experiment Details*

1) *Libraries and Frameworks*. For the video processing portion of our code, we used PyTorch as our deep learning model framework. We used OpenCV-Python to read and write video streams.

2) *Models*

a) *Traditional Detection & Classification Weighted Loss Yolov3 batch=8 width=512 height=512 channels=3 momentum=0.9 decay=0.0005 angle=0 saturation = 1.5 exposure = 1.5 hue=.1*

b) *Learning rate=0.001 burn in=100 max batches = 5000 policy=steps steps=4000,4500 scales=.1,.1*

c) *Simple CNAPS few-shot detection Mobilenetv3 + simple CNAPS Yolo + simple CNAPS batch=64 width=32 height=32 channels=3 momentum=0.9 decay=0.0005 learning rate=0.0001*

d) *Method to select the best hyperparameter configuration: Manually Tuned*

e) *The exact number of training and evaluation runs: 200 epoch*

*C. Computing Infrastructure*

For traditional detection and classification task, we trained our models on a container with an Intel Xeon CPU E5-2695v3 with default frequency of 2.30GHz, a GTX 2080 Ti GPU (11GB Memory) and 52.8 GB RAM. The video processing portion of our code is run in a standard Google Colab environment with GPU acceleration.

## ACKNOWLEDGEMENT

Thanks for Olivia Sun for her advice on the content of this paper and her editing work.